\documentclass[runningheads]{llncs}
\usepackage{subfigure}
\usepackage{graphicx}
\usepackage{latexsym}
\usepackage{amsxtra} 
\usepackage{amssymb}
\usepackage{amsmath}
\usepackage{color}
\usepackage{multicol,multirow,scalefnt}
\usepackage{subfig}
\usepackage{svg}
\usepackage[Algorithm]{algorithm}
\usepackage{algpseudocode}
\usepackage{array}

\newcommand{\toolname}{EA4CNN}

\begin{document}

\title{Optimizing Convolutional Neural Networks for Embedded Systems by Means of Neuroevolution}
\titlerunning{Optimizing Convolutional Neural Networks}

\author{Filip Badan and Lukas Sekanina\orcidID{0000-0002-2693-9011}}
\authorrunning{F. Badan and L. Sekanina}

\institute{Brno University of Technology, Faculty of Information Technology\\IT4Innovations Centre of Excellence \\ 
Bo\v{z}et\v{e}chova 2, 612 66 Brno, Czech Republic\\
\email{badan.filip@gmail.com, sekanina@fit.vutbr.cz}}

\maketitle

\begin{abstract}
 Automated design methods for convolutional neural networks (CNNs) have recently been developed in order to increase the design productivity. We propose a neuroevolution method capable of evolving and optimizing CNNs with respect to the classification error and CNN complexity (expressed as the number of tunable CNN parameters), in which the inference phase can partly be executed using fixed point operations to further reduce power consumption. 
 %We also demonstrate that the proposed method is useful when the evolutionary algorithm is seeded with an already trained CNN. 
 Experimental results are obtained with TinyDNN framework and presented using two common image classification benchmark problems -- MNIST and CIFAR-10. 


\end{abstract}

\section{Introduction}

\emph{Deep neural networks} (DNNs) currently show an outstanding performance in challenging problems of image, speech and natural language processing as well as in many other applications of machine learning. The design of high-quality DNNs is a hard task even for experienced designers because the state of the art DNNs have large and complex structures with millions of tunable parameters~\cite{AlexNet,sze:pieee17}. Automated DNN design approaches, often referred to as the \emph{Neural Architecture Search} (NAS), that have recently been developed, provide networks comparable with DNNs created by human designers. 
%After focusing on single-objective automated design methods a few years ago, where the main goal is to minimize the DNN error~\cite{DeepNEAT,Suganuma:GECCO2017,largescale}, the current trend is to develop multi-objective design methods in which the error is optimized together with other DNN properties such as computation requirements~\cite{MONAS,DPP-Net}. 

This paper deals with automated design and optimization of \emph{convolutional neural networks} (CNN), a subclass of DNNs primarily utilized for image classification. Our objective is to design and optimize not only with respect to the classification error, but also with respect to hardware resources needed when the final (trained) CNN is implemented in an embedded system with limited resources. As energy-efficient machine learning is a highly desired technology, various \emph{approximate implementations} of CNNs have been introduced~\cite{Panda:dnn16,Hashemi:Reda:date:2017}.
%for machine learning applications running on battery powered devices and low-cost Internet of %Things (IoT) nodes
Contrasted to the existing neuroevolutionary approaches trying to minimize the classification error as much as possible and assuming that CNN is executed using floating point (FP) operations on a Graphical Processing Unit (GPU)~\cite{MONAS,DPP-Net}, our target is a highly optimized CNN whose major parts are executed with reduced precision in fixed point (FX) arithmetic operations. 

We propose~\toolname~(Evolutionary Algorithms for Convolutional Neural Networks) -- a neuroevolution platform capable of evolving and optimizing CNNs with respect to the classification error and model complexity (expressed as the number of tunable CNN parameters), in which the inference phase can partly be executed using FX operations. One of our goals is to  demonstrate that the proposed method is capable of reducing the number of parameters of an already trained CNN and, at the same time, providing good tradeoffs between the classification error and CNN complexity. Experimental results are obtained with TinyDNN framewrok~\cite{TinyDNN} and presented using two common benchmark problems -- the classification of MNIST and CIFAR-10 data sets.

\section{Related Work}

Image classification conducted by CNNs is the state of the approach in the image processing domain. CNNs usually contain from four to tens layers of different types~\cite{sze:pieee17}. \emph{Convolutional layers} are capable of extracting useful features from the input data. In these layers, each neuron is connected to a subset of inputs with the same spatial dimensions as the tunable kernels. The convolution is computed as
$y = b + \sum_{i}\sum_{j}\sum_{k}(\textbf{x}_{i,j,k} \cdot \textbf{w}_{i,j,k}),$, where $\textbf{x}$ is the input subset, $\textbf{w}$ is the convolution kernel and $b$ is a scalar bias. 
%They are implemented by means of a high dimensional convolution of the input data with filter(s) whose coefficients (weights) are tunable parameters of the CNN. 
\emph{Pooling layers} combine, e.g. by means of averaging, a set of input values into a small number of output values to reduce the network complexity. \emph{Fully connected} (FC) layers are composed of artificial neurons; each of them sums weighted input signals (coming from a previous layer) and produces a single output. Convolutional layers and fully connected layers are typically followed by non-linear \emph{activation functions} such as $\tanh (\cdot)$ or rectified linear units (ReLU). The structure of the network is defined by hyperparameters (e.g., the number of layers, filters etc.) and this structure also determines the number of tunable parameters (weights and neuron biases). Modern CNNs also utilize normalization layers, residual connections, dropout layers etc. (see~\cite{AlexNet,sze:pieee17}).

In the \emph{training phase}, the objective is to optimize the CNN parameters in order to minimize a given \emph{error metric}. The training is a time-consuming iterative procedure which is typically implemented with the standard FP number representation. A trained CNN is then used, for example, for classification, in which an input image (a set of pixels) is classified to one of several classes. This (feed-forward) procedure is called \emph{inference} and only this procedure is typically implemented in low power hardware CNN accelerators~\cite{sze:pieee17}.

In order to automatically design the architecture (hyperparameters) and the parameters of CNNs, machine learning as well as evolutionary approaches have been proposed. Evolutionary design of neural networks (the so-called \emph{neuroevolution}) that was introduced three decades ago~\cite{neat}, is now being extended for CNN design~\cite{DeepNEAT}. 
%Important problems to solve are how CNN is represented in the chromosome (the so called direct or indirect encoding) and how training (learning) is incorporated into the evolutionary algorithm. 
As both CNN training and evolutionary optimization are very computationally expensive methods, the key problem of the current neuroevolution research is to reduce computational requirements and provide competitive CNNs with respect to the  human-created CNNs. Most papers are focused on single-objective automated design methods, where the main goal is to minimize the classification error of CNN running on a GPU~\cite{DeepNEAT,Suganuma:GECCO2017,largescale}. Recent works have been focused on multi-objective approaches in which the error is optimized together with the computation requirements~\cite{MONAS,DPP-Net}, but again, for GPU-based platforms. Evolved CNNs are now competitive with human-created CNNs for some challenging data sets; for example, some evolved CNNs achieve a $95$\% accuracy on CIFAR-10 data set. Note that a CNN with more than 1 million parameters is required in order to reach this accuracy and its training can take days on a GPU cluster~\cite{largescale}.

Another research direction is focused on energy efficient (hardware) implementations of CNNs -- with the aim of deploying advanced machine learning methods to low power systems such as mobile devices and IoT nodes. The most popular approach is to introduce approximate computing techniques to CNNs and benefit from the fact that the applications utilizing CNNs are highly error resilient (i.e., a huge reduction in energy consumption can be obtained for an acceptable loss in accuracy)~\cite{Panda:dnn16}. Approximate implementations of CNNs are based on various techniques such as innovative hardware architectures of CNN accelerators, simplified data representation, pruning of less significant neurons, approximate arithmetic operations, approximate memory access, weight compression and ``in memory'' computing ~\cite{sze:pieee17,Panda:dnn16,Hashemi:Reda:date:2017}. For example, employing the FX operations has many advantages such as reduced (i) power consumption per arithmetic operation, (ii) memory capacity needed to store the weights and (iii) processor-memory data transfer time.

To best of our knowledge, there has been no research on fully automated design of approximate CNNs by means of neuroevolution. As this is a very computationally expensive approach, we will focus this initial study on automated approximation of middle-size CNNs. Our method is based on simplifying the CNN architecture and reducing the precision of arithmetic operations.

\section{CNN Design and Optimization with Neuroevolution}

The proposed \toolname~framework exploits an evolutionary algorithm (EA) and TinyDNN library for the design and optimization of CNN-based image classifiers. TinyDNN was chosen because it can easily be modified with respect to the requirements of EA. TinyDNN can, however, be replaced by another suitable CNN library because \toolname~provides a general interface between EA and CNN implementations.
%\toolname~has been developed with respect to the following principles reflecting our requirements and a good practice reported in the literature: 
\toolname~is able to optimize and approximate an existing CNN, but it can also evolve a new CNN from scratch. CNN parameters as well as hyperparameters are optimized together. 
%\item Candidate CNNs are encoded directly as a list of layers, where each layer has many tunable parameters. The layers and their parameters are subject of the evolution. Some layers (such as the activations, fully connected layers and softmax) are always present in the phenotype and cannot be removed by genetic operators.
%\item The weights are determined by means of evolution (EA) or learning (TinyDNN). The weights associated with particular layer(s) are directly copied to next generation to reduce the need of training. 
%\item In the EA, the population diversity is maintained by introducing a speciation mechanism.
%\toolname~optimizes the CNN accuracy (in FP or FX implementation) and the CNN cost (the number of tunable CNN parameters). 

\subsection{Evolutionary Algorithm}

Algorithm~\ref{alg:EA_2} presents the EA developed for the design and optimization of CNNs. The EA is initialized with existing or randomly generated CNNs (line 1 in Algorithm~\ref{alg:EA_2}) and runs for $G_{max}$ generations (line 3). It employs a two-member tournament selection (line 6) to determine the parents that later undergo crossover (line 7; with probability $p_c$; see Section~\ref{sec:genop}  for details) and mutation (line 8; with probability $p_m$, see Section~\ref{sec:genop}). All offspring are continuously stored to the $Q$ set (line 9) and
%. All offspring (including those who have been modified by neither crossover nor mutation) 
undergo a training process implemented in TinyDNN (line 11).

Every new population is composed of the individuals selected from the sets of parents ($P$) and offspring ($Q$). The replacement algorithm (line 14) uses a simple speciation mechanism based on the CNN age (Section~\ref{sec:genop}). To prevent the overfitting, the data set is divided into three parts -- training set $D_{train}$, test set $D_{test}$ and validation set $D_{val}$. During the evolution, candidate individuals are trained using $D_{train}$ (line 11), but their fitness score is determined  using $D_{eval}$ (lines 2 and 13). At the end of the evolution process, the best solution is evaluated on the validation set $D_{val}$ and this result is reported. 

\begin{algorithm}[h!]
	\caption{Neuroevolution}
	\begin{algorithmic}[1]
		%\Procedure {EA}{}
		\State $P$ = Create Initial Population; // randomly or using existing CNN
		\State Evaluate($P, D_{test}$) using TinyDNN; i = 0;
		\While{($i < G_{max})$}
		    \State $Q = \emptyset$; \hspace{3cm} // a set of offspring 
		    \While{($|P| \neq |Q|$)}
			  \State $(a, b)$ = Tournament Selection $(P)$;
			  \State $(a',b')$ = Crossover$(a,b, p_{cross})$;
			  \State $a''$ = Mutation$(a', p_{mut})$; $b''$ = Mutation$(b', p_{mut})$;
			  \State $Q = Q \cup \{a''\} \cup \{b''\}$;
			\EndWhile		      
              \State Run TinyDNN's Training Algorithm for all NNs in $Q$ with $D_{train}$;
              \State Update the Age counter for all NNs.
		      \State Evaluate($Q, D_{test}$) using TinyDNN;
  			  \State $P$ = Replacement With Speciation $(P, Q)$;
			  \State $i = i + 1$;
		\EndWhile
	\end{algorithmic}
	\label{alg:EA_2}
\end{algorithm}

\subsection{CNN Encoding}

A candidate CNN is represented in the chromosome as a variable-length list of layers with a header containing the chromosome identifier, the age and the learning rate. Two types of layers can occur in the chromosome:
(1) \emph{Convolutional layer} with hyperparameters: kernel size, number of filters, stride size and padding.
(2) \emph{Pooling layer} with hyperparameters: stride size, subsampling type and subsampling size.

\begin{figure}[h]
\centering
\includegraphics[width=0.7\textwidth]{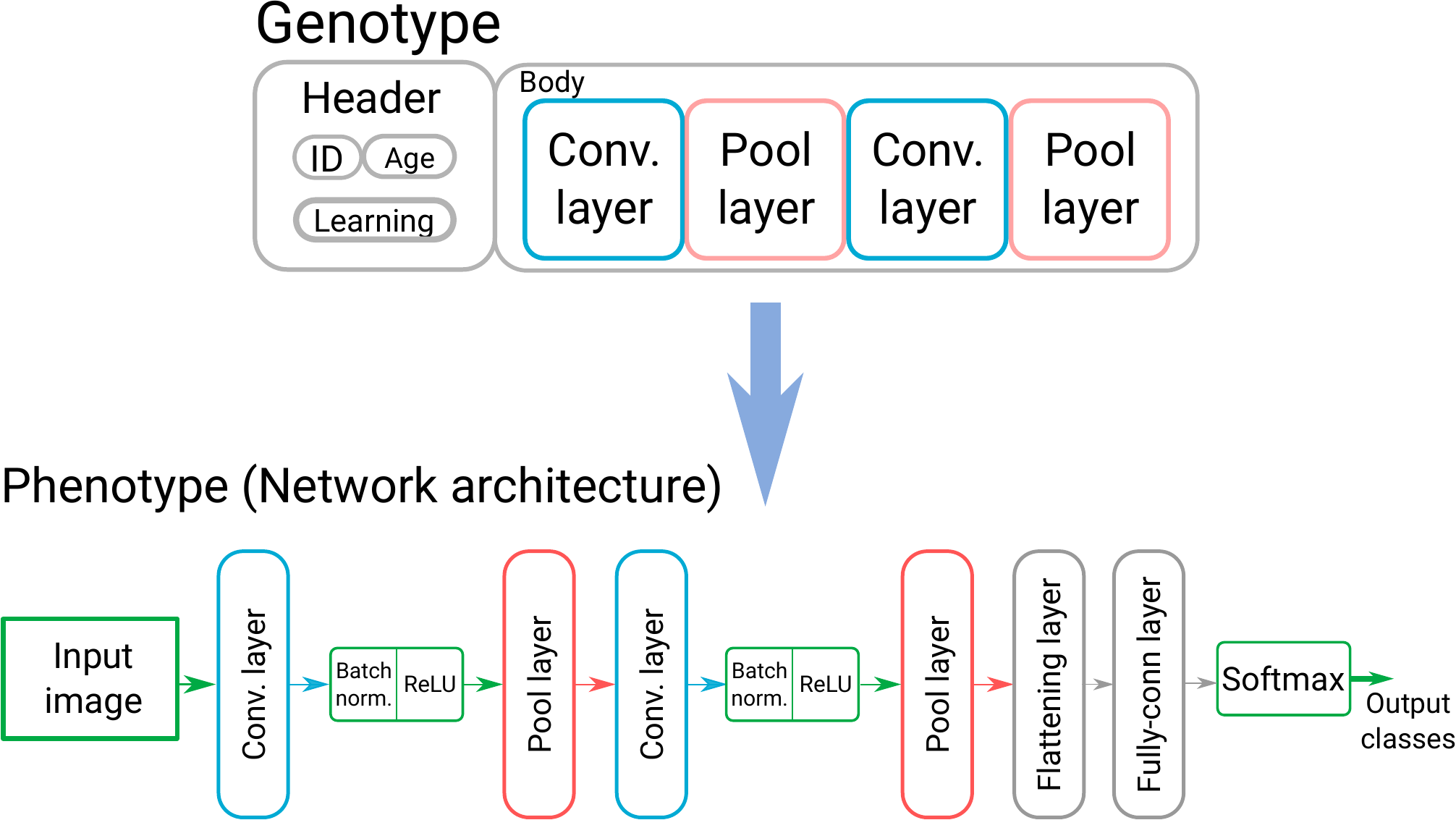}
\caption{Example of the genotype-phenotype mapping, where some parts of CNN (such as the flattening, fully connected and softmax layers) are not directly represented in the genotype.}
\label{fig:chromosome}
\end{figure}

Each convolutional layer is (obligatorily) followed by a batch normalization and ReLU activation. The last (obligatory) layers of each CNN are a convolutional flattening layer and a fully connected layer, followed by a softmax activation to obtain a classifier. These layers are not represented in the chromosome as shown in the example of genotype-phenotype mapping in Fig.~\ref{fig:chromosome}.

\subsection{Genetic Operators}
\label{sec:genop}

The mutation operator is applied with the probability $p_m$ per individual. One of the following mutation options (MO) is chosen with a predefined probability:
\begin{enumerate}
\item MO1: Weight reset -- all weights of a given layer are randomly generated.
\item MO2: Add a new layer -- a randomly generated layer (with randomly generated hyperparameters) is inserted on a randomly chosen position in CNN.
\item MO3: Remove layer -- one layer is removed from a randomly chosen position.
\item MO4: Modify layer -- some parameters of a randomly selected layer are randomly modified.
\item MO5: Modify hyperparameters of the fully connected layer -- the number of connections in the last fully connected layer is increased or decreased. 
\item MO6: Modify the learning rate (randomly).
\end{enumerate}

We use a simple one-point crossover operator on each pair of parents obtained with the tournament selection. 
If a CNN layer is modified by a genetic operator, \toolname~automatically ensures its correct connection to the previous/next layer. For example, superfluous weights are cut off or missing weights are added and randomly initialized.

\subsection{Training and Evaluation of Candidate CNNs}

As some candidate CNNs exist for many generations while others exist only for a short time, these long-lived  CNNs have more opportunities for a good training (line 11 in Algorithm~\ref{alg:EA_2}). It turns out that candidate CNNs do not have, in principle, the same chance during the selection and replacement process. Hence, inspired in~\cite{neat}, we introduced a speciation mechanism based on the \emph{network age}. A species is defined by all individuals having the same age. The age is increased with every new training process a given candidate CNN undergoes. We define $age_{max}$ as the maximum age a candidate network can obtain even if it undergoes more than $age_{max}$ training exercises. The reason for introducing this limit is to increase the selection pressure for networks that were trained many times. A typical setup of $age_{max}$ is $\sim|P|/2$. On the other hand, the network age is reset to the initial value if a given CNN is changed and its fitness is decreased as a consequence of crossover or mutation, e.g., after inserting or removing some layer(s) or changing parameters of the layer. The replacement is independently performed for all selected age levels; for example, if there are 5 age levels and the population size is 15 then 3 best-performing candidate CNNs are selected for each age level and copied to the new population. This algorithm is implemented by `Replacement With Speciation' on line 14 in Algorithm~\ref{alg:EA_2}. 

%In order to reduce computational resources, training of candidate solutions is simplified. 
For the new candidate individuals that are created by mutation or crossover, the principles of \emph{weight inheritance} are applied~\cite{largescale}. All the weights that can be reused in the offspring are copied from the parent(s) to the offspring. If needed, superfluous weights are cut off or missing weights are added and randomly initialized. 
Before each training phase is executed, $D_{train}$ is randomly shuffled~\cite{AlexNet}.

\subsection{Fitness Function}

The fitness function is based on the CNN accuracy ($a$ is the number correctly classified inputs divided by all inputs from the test set $D_{test}$) and the CNN relative size ($s_{rel}$ is the number of parameters divided by the number of parameters of the best CNN of the initial population):	
\begin{equation}
f = 
\begin{cases}
a * (k~* \frac{1}{\log (s_{rel} + 1)} + 1) & \text{if } a \ge a_{min}  \\
0 & \text{otherwise,} \\
\end{cases}
\label{eq:fitness}
\end{equation} 
where $k$ is a coefficient reflecting the impact of CNN size on the final fitness score and $a_{min}$ is the minimal acceptable accuracy. It is important to introduce $a_{min}$ as less complex CNNs providing unacceptable (low) classification accuracy would dominate the entire population. 

\subsection{Data Type and CNN Size Optimization}
\label{sec:datatypes}

Almost all major CNN design frameworks operate over (32 bit) FP numbers and their computation is optimized for arithmetic FP operations and accelerated using GPUs. In order to enable FX operations (in particular, FX multiplications conducted during the inference phase in convolutional and fully connected layers), we modified relevant parts of TinyDNN source code. When a multiplication has to be executed in these layers, the FP operands are converted to a given FX number format, the multiplication is performed in FX and the product is converted back to FP. While this process emulates the error introduced by FX representation in a low cost hardware, all the remaining CNN steps can be implemented with the (highly optimized) FP operations. Unfortunately, this implementation slows down the CNN simulations approx. 8 times in our case. 

When a CNN which should (partly) operate in the FX representation is evolved, we apply the aforementioned procedure in the fitness function (line 13 in Algorithm~\ref{alg:EA_2}); however, the training is completely conducted in FP.

In our study, a (signed) FX number is implemented using 16 bits, in which 8 bits are fractional. If a 32 bit FP multiplication is replaced with a 16 bit FX multiplication, energy consumption of this operation is reduced approx. $c_1 = 2.4$ times (for 65 nm technology~\cite{Hashemi:Reda:date:2017}). 
Let $E_{mult}$ denote the energy consumed by all multiplications performed during one inference phase carried out in a CNN embedded accelerator ($E_{mult}$ is approx. 20\,\% -- 40\,\% of the total energy required by the accelerator~\cite{sze:pieee17}). If the number of parameters of CNN is reduced from $par_{orig}$ to $par_{red}$ by \toolname~and 16 bit FX instead of 32 bit FP  multipliers are employed, $E_{mult}$ is reduced approx. $c_{1} \times par_{orig}/par_{red}$ times because each parameter is associated with at least one multiplication in CNN.  

\section{Experimental Setup}
\label{sec:setup}

\toolname~is implemented in C++. We utilized the parallel training of CNNs supported in TinyDNN (by means of OpenMP and SSE instructions). Experiments were executed on a computer node containing two Intel Xeon E5-2680v3 processors @ 2.5 GHz, 128 GB RAM and 24 threads. As the entire neuroevolution process is very time consuming (an average run in which 750 candidate CNNs are evaluated takes almost 72 hours for CIFAR-10), we typically generated only 50 populations of 15 individuals and performed only five independent runs for a particular setup. Hence, most EA parameters and CNN (hyper)parameters were set up on the basis of preliminary results from several test runs. 

\toolname~was evaluated using MNIST (10 digit classes) and CIFAR-10 (10 image classes) classification problems. 
MNIST consists of $28 \times 28$ pixel grayscale images of handwritten digits and includes 60\,000 training images and 10\,000 test images.
In CIFAR-10, the numbers of training and test images are 50\,000 and 10\,000, respectively, and the size of images is $32\times32$ pixels. For our purposes, these data sets were divided into three parts in such a way that there are 75\,\% vectors in $D_{train}$, 10\,\% vectors in  $D_{test}$ and and 15\,\% in  $D_{val}$. 

The basic setup of EA parameters is as follows: $G_{max} = 20 - 50$, $|P| = 8 - 15$, $p_{cross} = 0.35$, $p_{mut} = 0.7$, $age_{max} = |P|/2$, $k=0.5$, $a_{min} = 0.80$ for MNIST and 0.60 for CIFAR-10.
Mutation operators MO1 -- MO6 are used with the probabilities 0.41, 0.07, 0.03, 0.29, 0.10, and 0.10, respectively. 

Table~\ref{tab:CNN_params} summarizes the initial CNN hyperparameters for both data sets. Randomly generated networks of the initial populations contain from 1 to 8 layers in which all weights are randomly initialized to the close to zero values. TinyDNN utilizes the stochastic gradient descent learning method.

%The CNN tunable parameters that are used to estimate the network size are the number of weights and biases in convolutional and fully connected layers.

%In order to tune some of the parameters we performed additional experiments that will be reported in Section~\ref{sec:results}.
%\vspace{-0.5cm}
\begin{table}[t]
	\centering
	\caption{The initial setting of CNN hyperparameters in~\toolname. The hyperparameters given in the first part of the table can be modified during the evolution.}
	\label{tab:CNN_params}
	\tabcolsep 4pt 
	\begin{tabular}{l|c|c}
		\hline
		{Parameter/Data set}    & MNIST & CIFAR-10       \\ 
		\hline\hline
		Learning rate          & 0.1 & 0.1   \\
		Initial number of neurons in FC layers & 50 & 70   \\ 
		\hline
		Max. filters in a newly added layer & 12 & 20 \\ 
     	Max. pooling layer size & 4  & 4 \\ 
		Batch size & 32 & 32  \\ 
		Epochs for training & 1 & 1 \\ 
	\end{tabular}
\end{table}
%\vspace{-1cm}

\section{Results}
\label{sec:results}

%Experimental results reported in this section deal with a basic evaluation and setup of~\toolname~when all CNN operations are performed with FP number representation (Section~\ref{sec:res:basic}) and evolutionary approximation of existing CNNs (Section~\ref{sec:res:approx}).  

\subsection{Basic Evaluation of~\toolname}
\label{sec:res:basic}
\begin{figure}
	\centering
	{\includegraphics[width=0.46\textwidth]{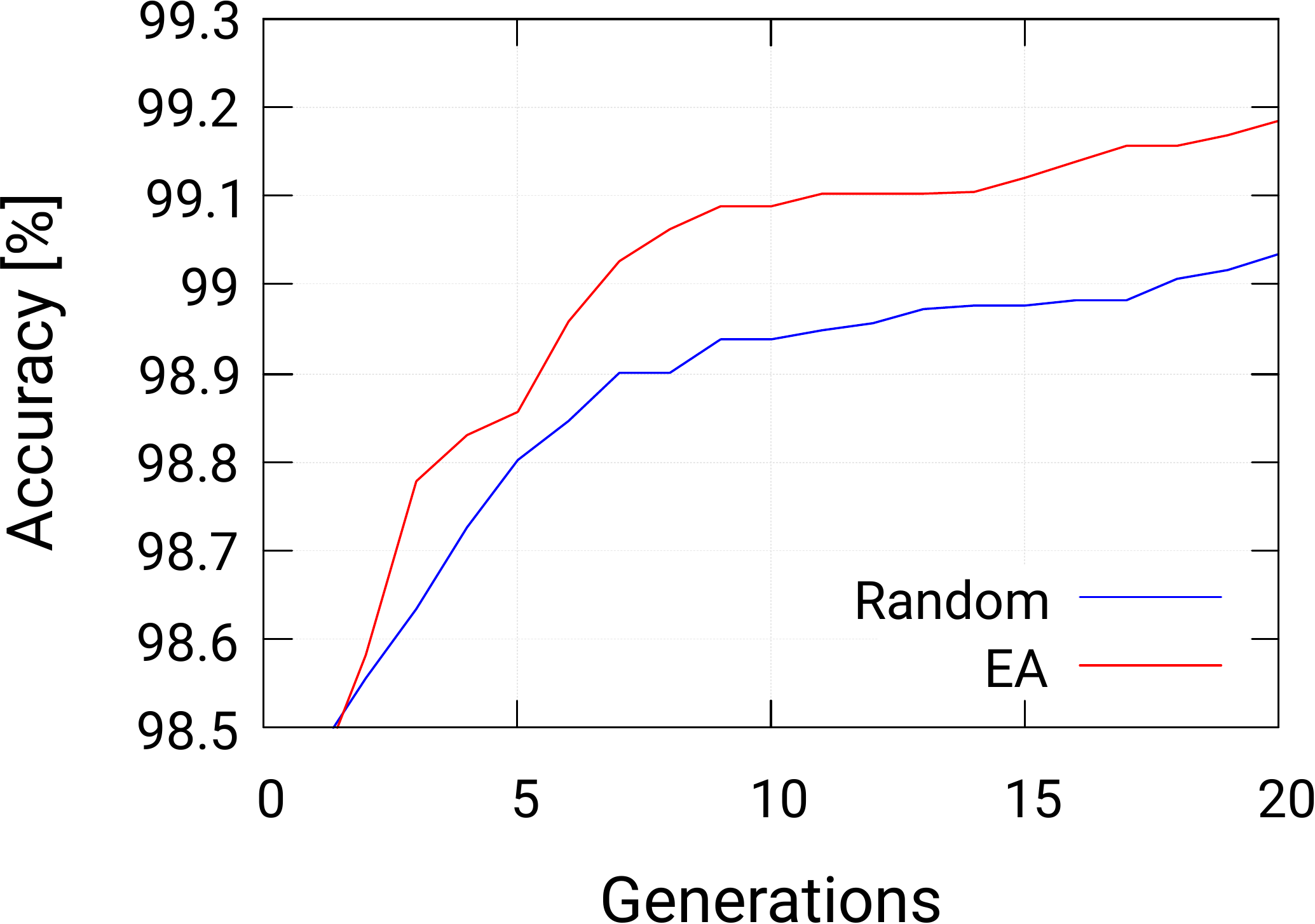}\label{plot:random_mnist}}
	\hfill
	{\includegraphics[width=0.46\textwidth]{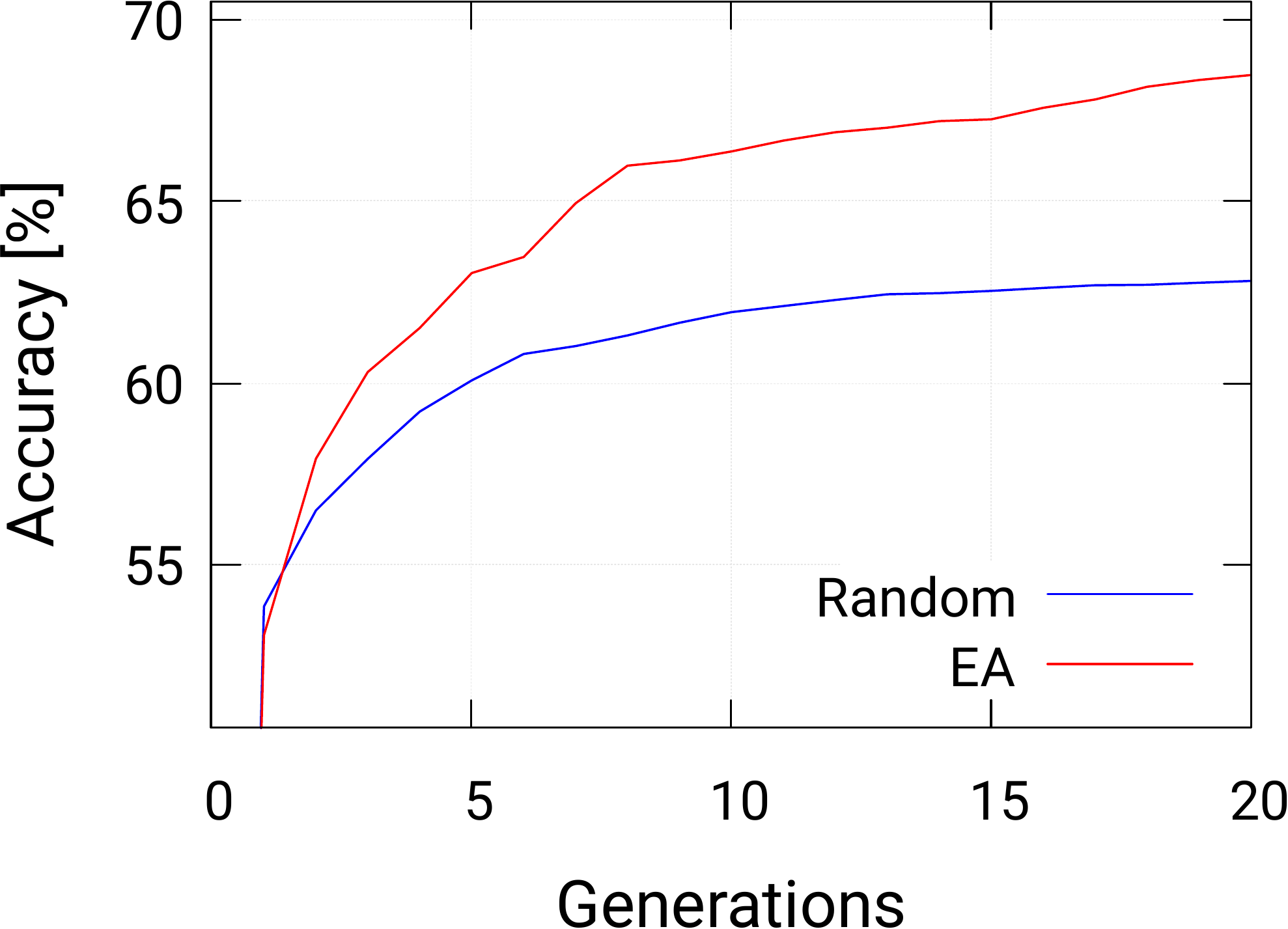}\label{plot:random_cifar}}
	\caption{The average accuracy obtained from five EA and five RS-CNN runs for MNIST (left) and CIFAR-10 (right) data sets.}
	\label{fig:ga:random}
\end{figure}

In the first experiment, we compared the randomly-initialized EA with a random search of CNNs (RS-CNN). EA used the setup presented in Section~\ref{sec:setup}, but $G_{max} = 20$, $|P|=8$ for MNIST and $|P|=12$ for CIFAR-10. RS-CNN starts with $|P|$ randomly generated CNNs and performs their training for $G_{max}$ epochs to ensure the same number of training exercises as in EA in which only one epoch of training is conducted for each candidate CNN in each generation. The average accuracy out of 5 independent runs of both algorithms is given in Fig~\ref{fig:ga:random}. Because MNIST classification is currently considered as a simple problem for NNs (the best reported accuracy is 99.79\%~\cite{sze:pieee17}), even randomly generated CNN architectures provide (after their training) almost perfect classification accuracy. The average number of parameters of resulting CNNs is 200k for RS-CNN, but only 58k for EA which indicates that EA can optimize not only accuracy but also the CNN complexity (resulting CNNs have only 1--2 convolutional layers). While EA is only slightly better than RS-CNN for MNIST, the difference in the average accuracy on CIFAR-10 is relatively high (5.7\% for 5 runs) which indicates that EA can also effectively increase the CNN size to improve the accuracy.

\begin{figure}
	\centering
	{\includegraphics[width=0.47\textwidth]{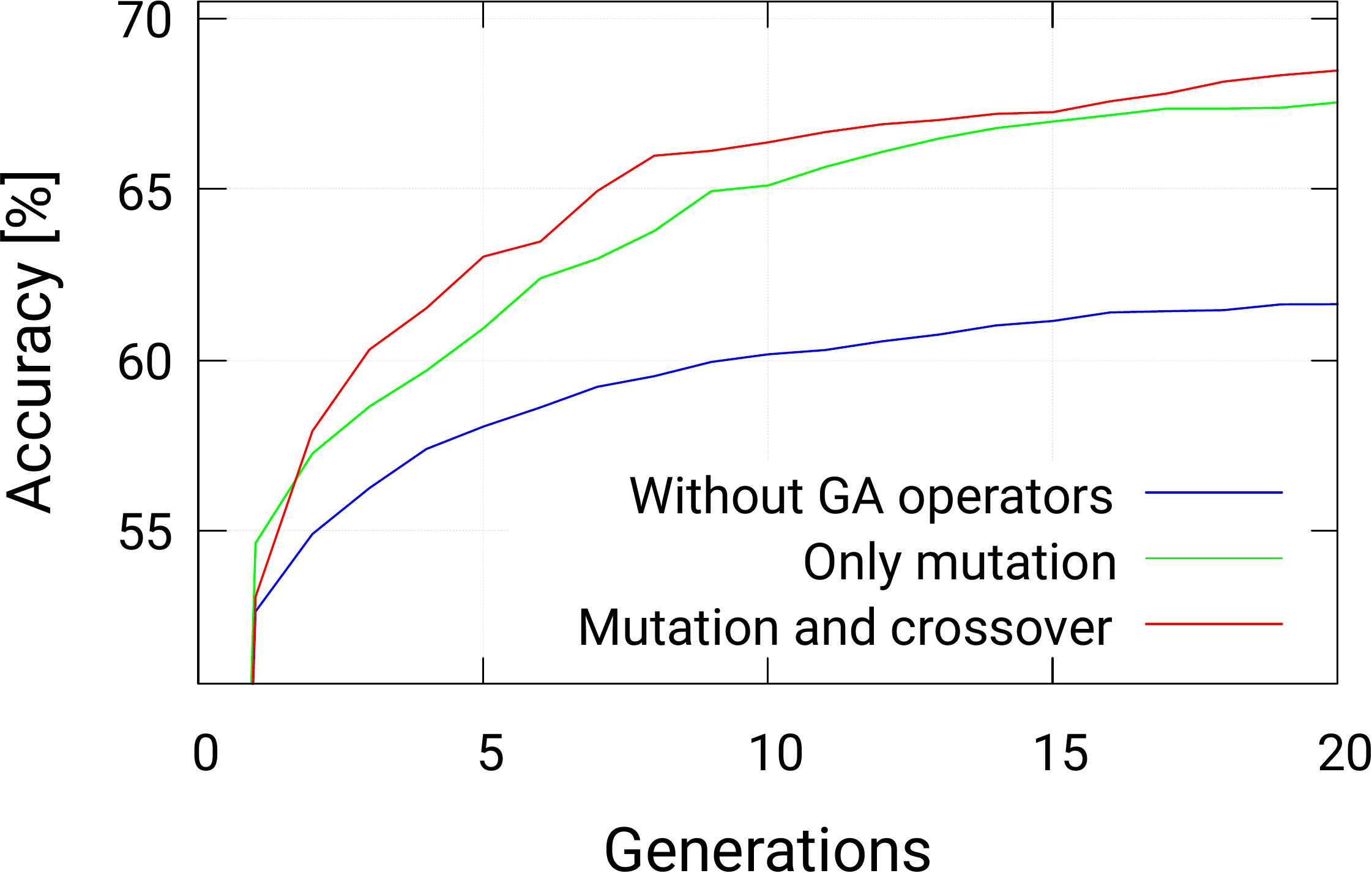}}
         {\includegraphics[width=0.47\textwidth]{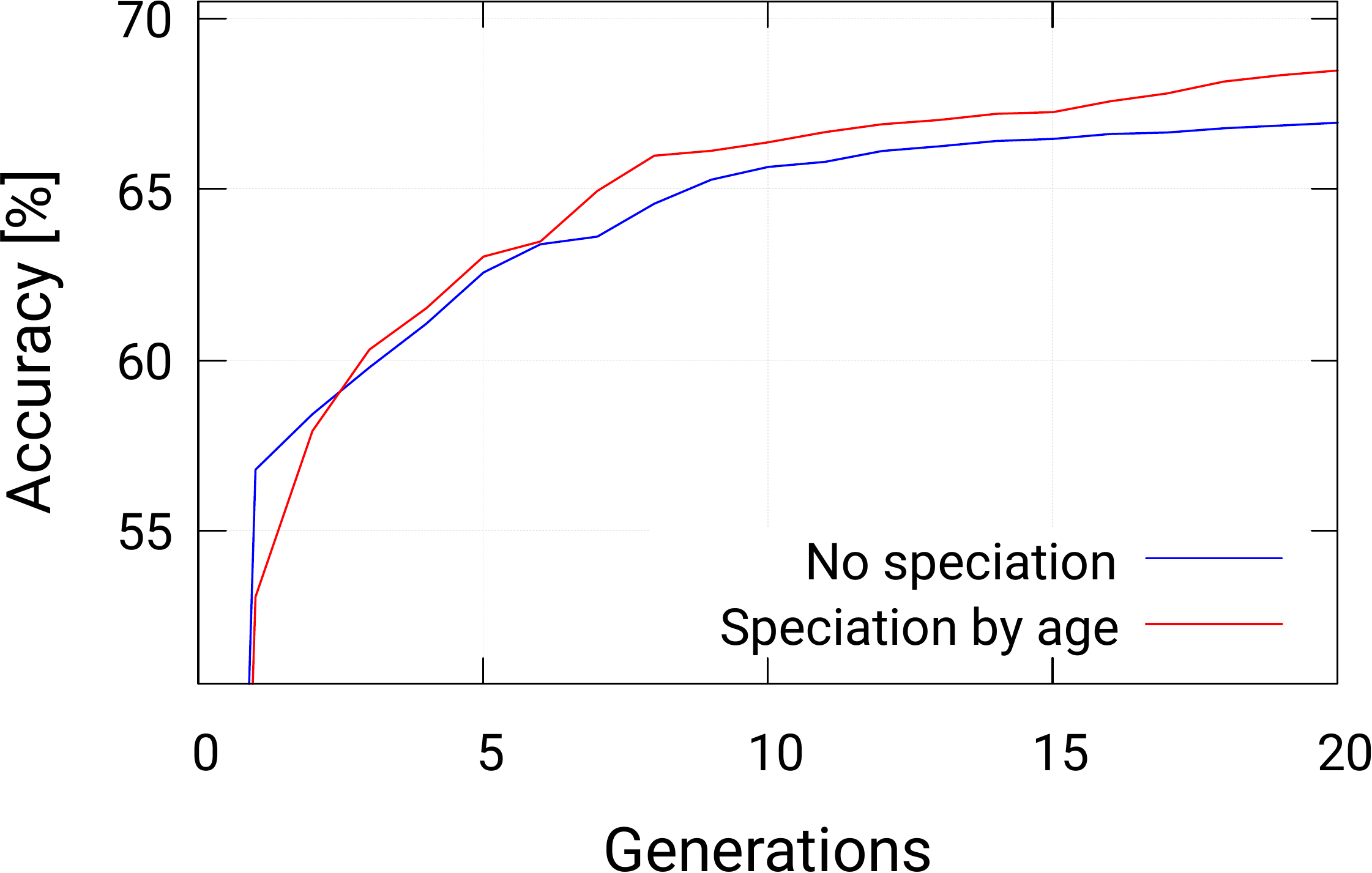}}
	\caption{Left: The average classification accuracy if EA uses selection only (EA1); selection and mutation (EA2); selection, mutation and crossover (EA3). Right: The average accuracy for EA3 with and without speciation (on CIFAR-10).}
	\label{fig:ga:genops}
\end{figure}

In the second experiment, we investigated the impact of genetic operators on the progress of evolution of CNNs. Let EA1 denote Algorithm~\ref{alg:EA_2} in which neither crossover nor mutation are used. EA1, in fact, does not introduce any new CNN structures, but optimizes how CNNs (randomly generated in the initial population) are selected for training by means of TinyDNN. Note that $D_{train}$ is randomly shuffled before each training. Higher-scored CNNs can thus undergo more training exercises and improve their fitness score. Let EA2 and EA3 denote EA1 with mutation ($p_{mut}=0.80$; no crossover) and EA1 with mutation ($p_{mut}=0.50$) and crossover ($p_{cross}=0.35$). The other parameters remained as given in Section~\ref{sec:setup}. 
The average classification accuracy out of 5 independent runs of EA1, EA2 and EA3 is given in Fig~\ref{fig:ga:genops} (left). Because of limited space only results on CIFAR-10 are reported. One can observe that performance of EA1 is roughly similar with RS-CNN. Incorporating the mutation operator (EA2) and crossover (EA3) leads to a higher classification accuracy of resulting CNNs.

Finally, Fig~\ref{fig:ga:genops} (right) illustrates the impact of employing the speciation mechanism on the accuracy during the CNN evolution. If the speciation ``is not used'' vs. ``is used'', the classification accuracy of resulting five CNNs is between 
59.93\% -- 72.96\% vs. 62.02\% -- 73.05\%; the average accuracy is 66.93\% vs. 68.46\%; the average depth of the network is 3.6 vs. 4.6 layers and the average number of parameters is 114k vs. 173k. We can conclude that the EA benefits from the proposed speciation mechanism.

\vspace{-1cm}
\subsection{Evolution of Approximate CNNs}
\label{sec:res:approx}

\begin{figure}
	\includegraphics[width=0.95\textwidth]{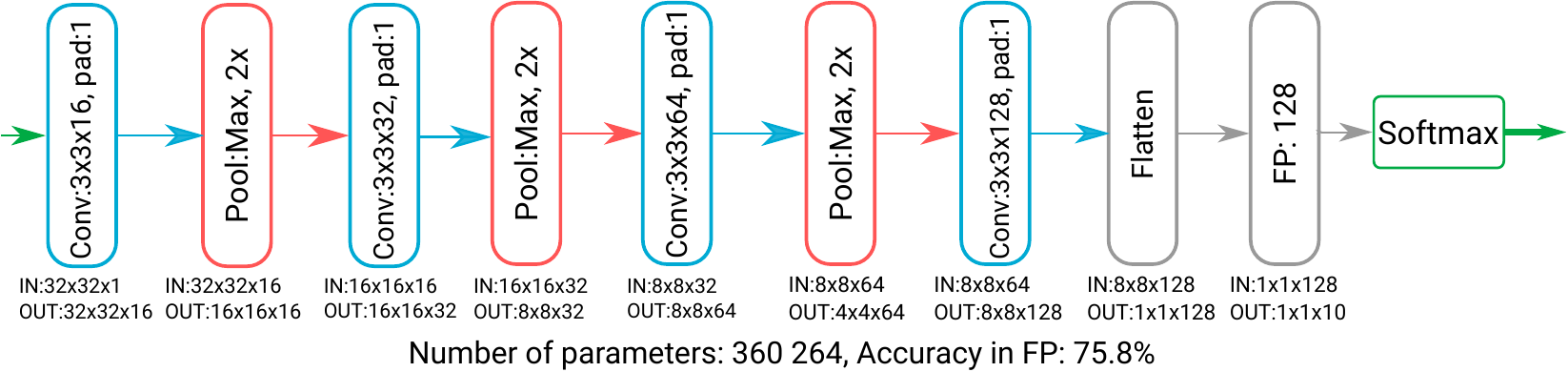}
	\includegraphics[width=0.95\textwidth]{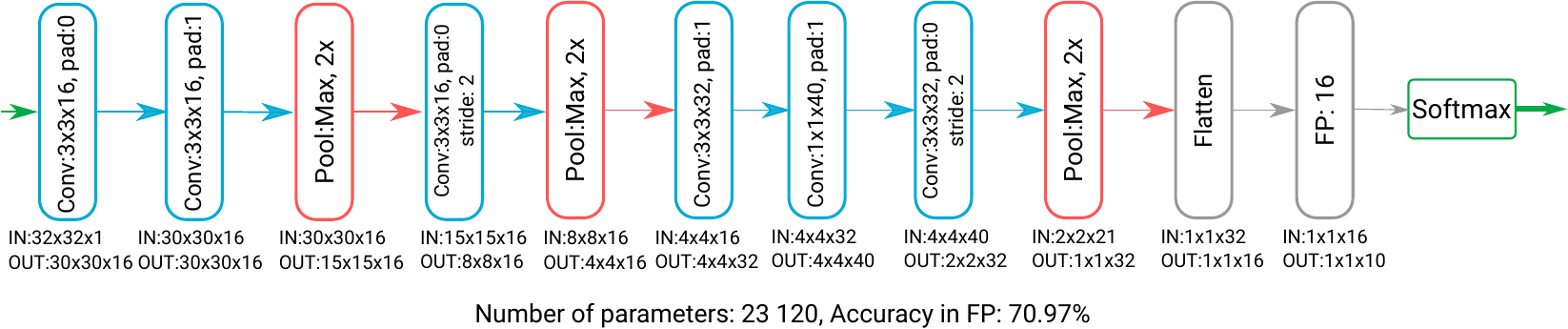}
	\caption{Hyperparameters and architecture of the baseline CNN (top) and one of the CNNs optimized with~\toolname~(bottom) for CIFAR-10 data set.}
	\label{figure:cifar_opt_both}
\end{figure}

The experiments reported in this section have started with a \emph{baseline CNN} shown in Fig.~\ref{figure:cifar_opt_both} (top) which contains 360,264 parameters, operates in FP and provides 75.8\% accuracy on CIFAR-10 (trained with TinyDNN). We decided to approximate this middle-size CNN as less complex CNNs are our target and our computational resources are limited.  

%\begin{figure}
%	\includegraphics[width=0.55\textwidth]{plots/cifar_start.pdf}
%	\includegraphics[width=0.60\textwidth]{plots/cifar_end.pdf}
%	\caption{Parameters and architecture of the baseline CNN (left) and one of the CNNs optimized with~\toolname~(right) for CIFAR-10 data set.}
%	\label{figure:cifar_opt_both}
%\end{figure}

\begin{figure}
    \centering
	\includegraphics[width=0.7\textwidth]{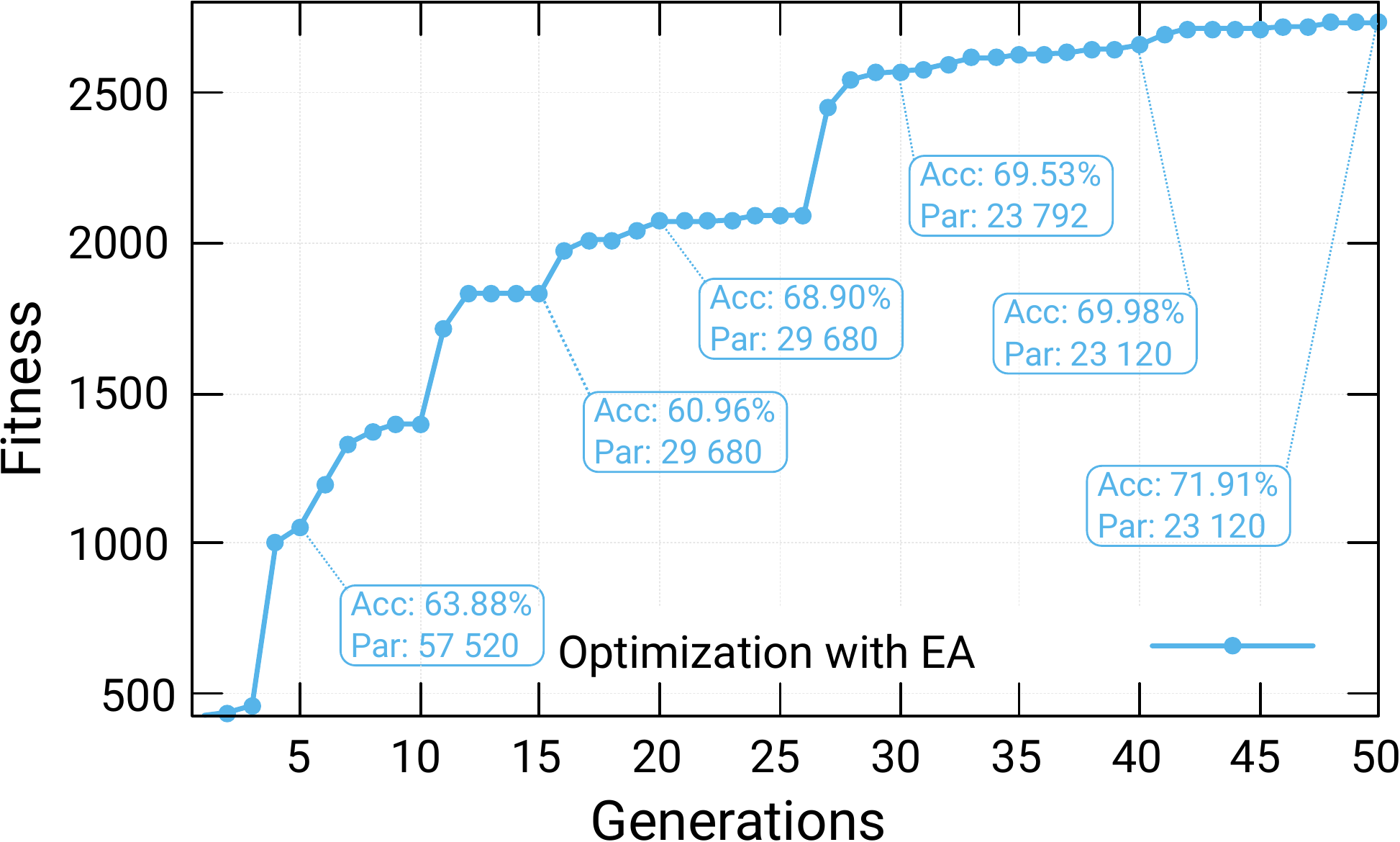}\label{plot:opt_fp_fitness}
	\caption{An example run of the evolutionary CNN approximation process on CIFAR-10 data set.}
	\label{fig:res:opt}
\end{figure}

Fig.~\ref{fig:res:opt} shows the fitness score, classification accuracy and complexity of CNNs obtained from a single run of the proposed EA which was seeded with the baseline CNN. The EA parameters are set according to Section~\ref{sec:setup}, but $k = 1$ to find good tradeoffs between the accuracy and the number of CNN parameters; $G_{max} = 50$, and $|P| = 15$ . Note that the accuracy shown in the plot is the test accuracy on $D_{test}$. The resulting CNN is presented in Fig.~\ref{figure:cifar_opt_both} (bottom). 
%All arithmetic operations were conducted in the FP number representation.

%\begin{figure}
%	\includegraphics[width=0.7\textwidth]{plots/cifar_opt.pdf}
%	\caption{Parameters and architecture of an optimized CNN.}
%	\label{figure:cifar_opt_end}
%\end{figure}

Table~\ref{tab:FP_res} summarizes the best tradeoffs obtained from multiple EA runs. CNN*-FP and CNN*-FX denote CNNs performing the multiplication operations in FP and FX representation, respectively. Because of limited computing resources, we could only execute 20 generations to evolve CNNs utilizing the FX representation, which negatively influenced the quality of CNN*-FX networks. For example, a similar classification accuracy ($\sim$ 67.5\%) was obtained by CNN2-FP and CNN1-FX, but CNN1-FX needs $2.9\times$ more parameters and hence it is less energy efficient despite the usage of FX multiplications.
Reduction in the energy ($E_{mult}$) needed for multiplication (calculated according to Section~\ref{sec:datatypes}, $c_1=2.4$) is clearly traded off for the loss in accuracy. \toolname~allowed us to obtain this reduction  by simplifying the CNN structure (fewer parameters) or/and employing FX operations. A more significant contribution of the FX representation is expected if \toolname~could prolong the optimization and thus further reduce the number of parameters in CNN*-FX networks. 

Table~\ref{tab:FP_res} also presents some CNNs that are available in the NAS literature. CNNs achieving a 90\% and higher classification accuracy on CIFAR-10 contain more than one million parameters~\cite{largescale} and their design takes days on a GPU, which is unreachable with our setup. 
%A CNN with 830 thousand parameters and showing 76.53\,\% accuracy was presented in in~\cite{Suganuma:GECCO2017}, but it was evolved in a simplified scenario with a very small training set. T
The impact of employing the FX representation was reported for a human-created CNN (based on AlexNET~\cite{AlexNet}) which exhibits 81.22\,\%, 79.77\,\% and 77.99\,\% accuracy in FP, 16 bit FX and 8 bit FX, respectively. The 16 bit and 8 bit FX implementations reduce the energy requirements approx. 2.5 and 6.8 times, respectively. Contrasted to our approach, these FX designs only implemented  the original FP implementation with reduced precision in FX, i.e. without optimizing the CNN architecture.

\vspace{-0.5cm}
\begin{table}
	\centering
	\caption{Examples of CNNs and their parameters obtained from the evolutionary approximation conducted with~\toolname~ and from literature (for CIFAR-10).}
	\label{tab:FP_res}
	\tabcolsep 3pt 
	\begin{tabular}{l|r|r|r|r}
	\hline
		CNN & {Parameters} & {Accuracy} & {Layers} & $E_{mult}$ reduction\\ 
	%	\hline
		\hline
		\multicolumn{5}{c}{Evolved with~\toolname}\\
		\hline
		Baseline CNN (FP) & 360,264 & 75.80\,\%  & 7   & $1.0~$\\
		\hline
		CNN1-FP & 8\,480 & 64.33\,\% & 9   & $42.9\times$ \\ 
		CNN2-FP & 12\,784 & 67.50\,\%  & 7 & $28.1\times$\\ 
     	CNN3-FP & 15\,728 & 68.92\,\%  & 8 & $22.9\times$\\
		CNN4-FP & 23\,120 & 70.97\,\%  & 9  & $15.6\times$\\
		CNN5-FP & 0.17 M & 72.96\,\%  & 6  & $2.1\times$\\
		\hline
		CNN1-FX (16 bit) & 36\,720 & 67.66\,\% & 11 & $23.6\times$\\
		CNN2-FX (16 bit) & 30\,672 & 66.52\,\%  & 8  & $28.2\times$\\
		CNN3-FX (16 bit) & 19\,632 & 65.63\,\% & 7 & $44.0\times$\\
		\hline
		\multicolumn{5}{c}{From literature}\\
		\hline
		\cite{Suganuma:GECCO2017} (FP) default scenario & 1.68 M & 94.02\,\% & -- & -- \\
		\cite{Suganuma:GECCO2017} (FP) small data set (5k) & 0.83 M & 76.53\,\% & --& --\\
		\cite{largescale} (FP) & 5.40 M & 94.60\,\% & -- & -- \\
		% One shot \cite{Wistuba:19} (FP) & 5.70 M & 97.92\,\% & -- & -- \\ % asi v soucasnosti nejlepsi
		\hline
		ALEX \cite{Hashemi:Reda:date:2017} (FP) & $\sim 10$ M & 81.22\,\% & -- & $1.0~{}$ \\
		ALEX \cite{Hashemi:Reda:date:2017} (FX, 16 bit) & $\sim 10$ M & 79.77\,\% & -- & $\sim 2.5\times$ \\
		ALEX \cite{Hashemi:Reda:date:2017} (FX, 8 bit) & $\sim 10$ M & 77.99\,\% & -- & $\sim 6.8\times$ \\
	\end{tabular}
\end{table}

\vspace{-0.5cm}
\section{Conclusions}

The proposed \toolname~platform can automatically evolve a CNN (with 29k parameters) showing almost the state-of-the-art accuracy (99.36\,\%) for the MNIST task.
Evolved CNNs for CIFAR-10 are far from the state-of-the-art, but it was expected because we used only the basic CNN techniques (no data augmentation, residual connections, dropout layers etc.) and very limited computing resources. However, we demonstrated that \toolname, if seeded with a trained CNN, can find interesting tradeoffs between the accuracy and implementation cost.

Our future work will focus on improving the search quality by incorporating advanced CNN techniques and employing more computing resources. 
%In addition to the automated architecture modification and reducing the precision, 
We will also explore more options for optimizing the CNN cost in order to develop a fully automated holistic CNN approximation method.

\section*{Acknowledgments} 
This work was supported by the Ministry of Education, Youth and Sports, under the INTER-COST project LTC~18053, NPU II project IT4Innovations excellence in science LQ1602 and by Large Infrastructures for Research, Experimental Development and Innovations project ``IT4Innovations National Supercomputing Center -- LM2015070''.

\bibliographystyle{splncs04}
\bibliography{main}

\end{document}